\documentclass{article}
\usepackage{spconf,amsmath,graphicx}
\usepackage{algorithm}
\usepackage{algorithmic}
\usepackage{multirow}
\usepackage{booktabs}
\usepackage{xcolor}

\usepackage{pifont}
\usepackage{romannum}
\usepackage{tablefootnote}

\newcommand{\cmark}{\ding{51}}%
\newcommand{\xmark}{\ding{55}}%
\DeclareMathOperator*{\argmax}{argmax}
\renewcommand{\paragraph}[1]{\noindent\textbf{#1}\quad}

\title{Towards Adversarially Robust Continual Learning}
\name{Tao Bai\sthanks{Work done during internship at Sony AI. 
}$^{1}$ \qquad Chen Chen$^{2}$ \qquad Lingjuan Lyu$^{2}$\sthanks{Corresponding Author.} \qquad Jun Zhao$^{3}$ \qquad Bihan Wen$^{1}$}
\address{
$^{1}$  School of Electrical and Electronic Engineering, Nanyang Technological University \\
$^{2}$ Sony AI \\
$^{3}$  School of Computer Science and Engineering, Nanyang Technological University \\
}

\begin{document}
\maketitle
\begin{abstract}
Recent studies show that models trained by continual learning can achieve the comparable performances as the standard supervised learning and the learning flexibility of continual learning models enables their wide applications in the real world.
Deep learning models, however, are shown to be vulnerable to adversarial attacks.
Though there are many studies on the model robustness in the context of standard supervised learning, protecting continual learning from adversarial attacks has not yet been investigated. 
To fill in this research gap, we are the first to
study adversarial robustness in continual learning and propose a novel method called \textbf{T}ask-\textbf{A}ware \textbf{B}oundary \textbf{A}ugmentation (TABA) to boost the robustness of continual learning models. 
With extensive experiments on CIFAR-10 and CIFAR-100, we show the efficacy of adversarial training and TABA in defending adversarial attacks.

\end{abstract}
\section{Introduction}
\label{sec:intro}
Continual learning studies the problem of learning from an infinite stream of data, with the goal of gradually extending acquired knowledge and using it for future learning~\cite{Delange_Aljundi_Masana_Parisot_Jia_Leonardis_Slabaugh_Tuytelaars_2021}. The major challenge is to learn without catastrophic forgetting: performance on a previously learned task or domain should not significantly degrade over time when new tasks or domains are added. To this end, researchers have proposed various methods~\cite{kirkpatrick2017overcoming, lee2017overcoming,rebuffiICaRLIncrementalClassifier2017, lopez2017gradient}
to reduce the computational costs while maintaining the performances for learned tasks.
As such, continual learning has %
made a wide range of real-world applications into reality recently~\cite{lee2020clinical, gupta2020neural}. 

Though the training process of continual learning is quite different from regular supervised learning, the model trained with continual learning is exactly the same as the regular supervised learning during inference. 
Recent studies~\cite{szegedy2013intriguing,goodfellow2014explaining} on adversarial examples reveal the vulnerabilities of well-trained deep learning models, which are easy to break through.
Thus, it's natural to assume that models trained with continual learning suffer from adversarial examples as well. 
Considering the real-world applications of continual learning models, it
is essential to protect continual learning models against adversarial attacks. 
There have been a number of studies exploring how to secure the deep learning models against adversarial examples~\cite{madryDeepLearningModels2018, bai2021recent}, but surprisingly, protecting continual learning from adversarial attacks has not been fully studied. 

\begin{figure}[t]
    \centering
    \includegraphics[width=\columnwidth]{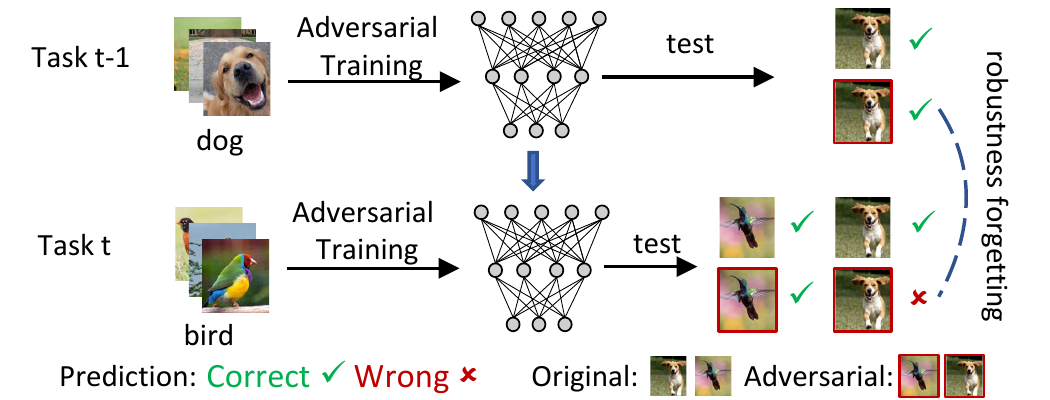}
    \caption{Robust continual learning and the issue of robustness forgetting. (\textit{e.g.} ``adversarial dog" is predicted wrong after training on Task t).}
    \label{fig:intro}
\end{figure}

To bridge the gap between continual learning and adversarial robustness, we focus on the replay-based continual learning methods and take the first step to develop robust continual learning methods. 
As we stated above, nevertheless, data from previously learned tasks in continual learning are partially accessible, causing the imbalance between previous tasks and new tasks.
In this case, models trained in current stage usually tend to overfit the new class data.
As such, the catastrophic forgetting of adversarial robustness is inevitable in robust continual learning, \textit{i.e.} when leveraging adversarial training~\cite{madryDeepLearningModels2018,bai2021recent} in continual learning for adversarial robustness (see Fig.~\ref{fig:intro}) 
Preventing forgetting, or in other words, preserving learned knowledge refers to maintaining the previously learned decision boundary among classes~\cite{zhuClassIncrementalLearningDual2021}.
We thus propose a novel approach called \textbf{T}ask-\textbf{A}ware \textbf{B}oundary \textbf{A}ugmentation (TABA) for maintaining the decision boundaries for adversarial training in continual learning settings. 

Our contributions are summarized as follows:
\begin{enumerate}
    \item To the best of our knowledge, we are the first to investigate the security issues in continual learning %
    and improving the adversarial robustness by leveraging adversarial training. 
    \item We further identify the catastrophic forgetting of adversarial robustness and propose a novel approach called \textbf{T}ask-\textbf{A}ware \textbf{B}oundary \textbf{A}ugmentation (TABA) for enhancing adversarial training and continual learning.
    \item With experiments on popular datasets like CIFAR10 and CIFAR100, we show the efficacy of TABA in different continual learning scenarios.
\end{enumerate}

\section{Related Works}
\label{sec:related}
Continual learning is widely studied in the last few years, which assumes data comes in a sequential way~\cite{kirkpatrick2017overcoming, lee2017overcoming,dong2022federated}.
There are, however, only a few works studying the security issues in continual learning~\cite{khan2022susceptibility,guo2020attacking}.
It is empirically shown the importance of robust features in continual learning~\cite{9892970}.
Authors of \cite{9937385} proposed to incorporate adversarial training with continual learning to enhance the robustness, as adversarial training has been validated in other deep learning tasks~\cite{madry2018towards,pmlr-v97-zhang19p,bai2021recent}.
In this paper, we study how to leverage adversarial training in continual learning and alleviate the catastrophic forgetting of adversarial robustness.

\section{Approach}
\label{sec:approach}
\subsection{Problem Definition}
In this work, we focus on the robust multi-class classification problem, which involves the sequential learning of $\mathcal{T}$ stages/tasks consisting of disjoint class sets.
Formally, at learning stage $t \in \{2, \dots \mathcal{T}\}$, given a model trained on an old dataset $\mathcal{X}_o^{t-1}$ from stage $\{1, \dots t-1\}$, our goal is to learn a unified classifier for both old classes $\mathcal{C}_o$ and new classes $\mathcal{C}_n$.
The training data at stage $t$ is denoted as $\mathcal{X}^t=\mathcal{X}_n^t \cup \tilde{\mathcal{X}}_o^{t-1}$, where $\tilde{\mathcal{X}}_o^{t-1}$ is a tiny subset of $\mathcal{X}_o^{t-1}$.
Thus, the challenge in continual learning is retraining the original model with the severely imbalanced $\mathcal{X}^t$ to boost the robustness on all seen classes while avoiding catastrophic forgetting. 

\subsection{Revisiting Distillation for Catastrophic Forgetting }
Knowledge distillation~\cite{hinton2015distilling} is firstly introduced to continual learning by Learning without forgetting (LwF)~\cite{li2017learning} and adapted by iCaRL~\cite{rebuffiICaRLIncrementalClassifier2017} for the \textit{multi-class} continual learning problem. 
Typically, the loss function of such distillation-based methods consists of two terms for each training sample $x$: the classification loss $\mathcal{L}_{ce}$ and the distillation loss $\mathcal{L}_{dis}$.
Specifically, the classification loss $\mathcal{L}_{ce}$ is expressed as 
\begin{equation} 
\small
\mathcal{L}_{ce}(x)=-\sum_{i=1}^{|\mathcal{C}|} y_i \log \left(p_i\right),
\label{eqn:loss_ce}
\end{equation}
where $\mathcal{C} = \mathcal{C}_o \cup \mathcal{C}_n$, $y_i$ is the $i_{th}$ value of the one-hot ground truth $y$, and $p_i$ is the $i_{th}$ value of predicted class probability $p$.
The goal of $\mathcal{L}_{dis}$ is to preserve knowledge obtained from previous data, which is expressed as
\begin{equation}
\small
\mathcal{L}_{dis}(x)=-\sum_{i=1}^{\left|\mathcal{C}_{\mathrm{o}}\right|} \left(p^*\right) \log \left(p\right),
\label{eqn:loss_dis}
\end{equation}
where $p^*$ is the soft label of $x$ generated by the old model.
It, however, is observed in ~\cite{rebuffiICaRLIncrementalClassifier2017} that there is tendency of classifying test samples to new classes by LwF.
Thus, iCaRL utilized \textit{herd selection} to better approximate the class mean vector of old classes, where samples that are close to the center of old classes are selected.

Recall that our goal is to obtain a robust model trained in the continual learning manner.
To gain robustness, adversarial training is inevitable, which requires augmenting datasets with adversarial examples in every training iteration.
Following the definition of continual learning, we can derive the loss function of \textbf{R}obust \textbf{C}ontinual \textbf{L}earning~(RCL).
With adversarial training, we should replace the input $x$ in Equation~(\ref{eqn:loss_ce}) and~(\ref{eqn:loss_dis}) with its adversarial counterpart $x_{adv}$, which is solved by 
\begin{equation}
\small
x_{adv} = \argmax_{||x_{adv}-x||_p \leq \epsilon} (\mathcal{L}_{ce}(x_{adv})),
\label{eqn:adv example}
\end{equation}
where $\epsilon$ is the allowed magnitude of perturbations in $p$-norm.
Thus, the loss function of robust continual learning would be 
\begin{equation}
\small
\mathcal{L}_{RCL} = \mathcal{L}_{ce}(x_{adv}) + \mathcal{L}_{dis}(x_{adv})
\end{equation}

Nevertheless, simply combining adversarial training with continual learning is not enough.
From the perspective of adversarial training, centered exemplars are not helpful for the forgetting of adversarial robustness. 
Recent studies~\cite{zhangGeometryawareInstancereweightedAdversarial2020,chenDecisionBoundaryawareData2022} pointed out that not all data points contribute equally during adversarial training and samples that are close to the decision boundaries should be emphasised. 
Therefore, how to deal with the exemplar set during adversarial training is essential for robust continual learning.  
In addition, adversarial training is more data-hungry than standard training.
The significant imbalance between old classes and new classes can be more severe. 
In this work, we aim to tackle these problems by incorporating data augmentation with adversarial training.

\subsection{Task-Aware Boundary Augmentation}
Preventing catastrophic forgetting of adversarial robustness in continual learning is equivalent to maintaining the decision boundary learned by adversarial training.
One direct way to do so is to introduce some samples close to the decision boundaries to the exemplar set (named Boundary Exemplar in Section~\ref{sec:exp} and Table~\ref{tab:setting1}).
However, this makes the exemplar selection process more sophisticated because the ratio of centered samples and boundary samples is hard to decide.
In addition, such mixed exemplar set may have negative influence on the approximation of old classes, which may downgrade the model performance.
Another potential solution is Mixup~\cite{zhang2018mixup}, where the dataset is augmented by interpolating different samples linearly.
Mixup, however, is not specially designed for adversarial training or continual learning.
It breaks the local invariance of adversarially trained models by linear interpolation and worsens the imbalance between old tasks and new tasks.

Inspired by Mixup, we propose \textbf{Task-Aware Boundary Augmentation~(TABA)} to augment the training data $\mathcal{X}$ by synthesizing more boundary data, which can be plugged in RCL easily.
Compared to Mixup, TABA is specially designed for adversarial training and continual learning.
The differences are summarized as below.
\textit{First}, TABA doesn't select samples in the whole dataset but from the boundary data.
The reason is that boundary data is easier to attack and contributes more to adversarial robustness~\cite{zhangGeometryawareInstancereweightedAdversarial2020}.  
We can obtain the boundary data for free because adversarial training requires generating adversarial examples.
Misclassified samples in the previous iteration are marked as the boundary data, which is denoted by $\mathcal{B}$.  
\textit{Second}, to deal with the data imbalance issue in continual learning, TABA selects samples from two sets: one is boundary data from $\tilde{\mathcal{X}}_o^{t}$ and the other is boundary data from $\mathcal{X}_n^t$, denoted as $\mathcal{B}_o$ and $\mathcal{B}_n$, respectively.
In this way, the augmented data can help maintain the learned decision boundaries in the previous stage.
\textit{Third}, we restrict the interpolation weight $\lambda$ to a interval of $[0.45, 0.55]$ rather than $[0,1]$ in Mixup to avoid the linearity, which is decided empirically.
The augmented samples can also be closer to the decision boundaries, compared to samples provided by Mixup.%

The augmented sample $(\bar{x}, \bar{y})$ by our TABA can be defined as follows:
\begin{equation}
\small
\begin{aligned}
&\bar{x}=\lambda x_o+(1-\lambda) x_n \\
&\bar{y}=\lambda y_o+(1-\lambda) y_n,
\end{aligned}
\label{eqn:taba}
\end{equation}
where $\lambda$ is the interpolation weight, $(x_o, y_o) \in \mathcal{B}_o$  and $(x_n, y_n) \in \mathcal{B}_n$.

Accordingly, the final loss function of RCL with TABA (RCL-TABA) would be 
\begin{equation}
\centering
\small
\begin{aligned}
&\mathcal{L}_{final} = \mathcal{L}_{TABA} + \mathcal{L}_{RCL} \\
&\mathcal{L}_{TABA} = \mathcal{L}_{ce}(\bar{x}_{adv}) + \mathcal{L}_{dis}(\bar{x}_{adv}).
\end{aligned}
\label{eqn:final}
\end{equation}
The training details of RCL-TABA are in Algorithm~\ref{alg1}.

\begin{algorithm}[t]
	\renewcommand{\algorithmicrequire}{\textbf{Input:}}
	\renewcommand{\algorithmicensure}{\textbf{Output:}}
	\caption{Robust continual learning with task-aware boundary augmentation (RCL-TABA)}
	\label{alg1}
	\begin{algorithmic}[1]
	    \STATE Randomly initialize model $f^0$, old task data $\tilde{\mathcal{X}}_{o}^{0} = \O$
	    \FOR{$t = \{1, \dots \mathcal{T}\}$}
	        \STATE \textbf{Input:} model $f^{t-1}$, new task data $\mathcal{X}_{n}^t$, training epochs $E$, number of batches $M$, original batch size $m$, interpolation batch size $m'$
	        \STATE \textbf{Output:} model $f^{t}$
	        \STATE $f^{t} \xleftarrow{} f^{t-1}$, $\mathcal{X}^t = \mathcal{X}_n^t \cup \tilde{\mathcal{X}}_o^{t-1}$, $\mathcal{B}_0 = \mathcal{X}^t$
	        \FOR{ $e = \{1, \dots E\}$}
	        \STATE  $\mathcal{B}_e = \O$
	        \STATE Compute augmentation set $\bar{\mathcal{X}^t}$ from $\mathcal{B}_{e-1}$ by Eq.~(\ref{eqn:taba})
	        \FOR{$mini-batch = \{1, \dots M\}$}
    	        \STATE Randomly sample $\{(x_i, y_i)\}_{i=1}^m$ from $\mathcal{X}$
    	        \FOR{$i = \{1, \dots m\}$} 
        	        \STATE  Generate adversarial data $x_i^{adv}$ by Eq.~(\ref{eqn:adv example}) 
        	        \IF{$f(x_i^{adv}) \neq y_i$ }
            	        \STATE $\mathcal{B}_e \xleftarrow{} \mathcal{B}_e \cup {(x_i, y_i)}$
        	        \ENDIF
    	        \ENDFOR
    	        \STATE Randomly sample $\{(\bar{x_i}, \bar{y_i})\}_{i=m+1}^{m+m'}$ from $\bar{\mathcal{X}}$
    	        \FOR{$i = \{m+1, \dots {m+m'}\}$} 
    	        \STATE Generate adversarial data $\bar{x_i}^{adv}$ by Eq.~(\ref{eqn:adv example}) 
    	        \ENDFOR
    	        \STATE optimize $f_t$ on $\{(\bar{x_i}, \bar{y_i})\}_{i=1}^{m+m'}$ by Eq.~(\ref{eqn:final})
	        \ENDFOR
	        \ENDFOR
	        \STATE update $\tilde{\mathcal{X}}_o^{t}$ by class using \textit{herd selection}~\cite{rebuffiICaRLIncrementalClassifier2017}
	    \ENDFOR
	\end{algorithmic}  
\vspace{-2pt}
\end{algorithm}

\section{Experiments}
\label{sec:exp}
\subsection{Settings}
\paragraph{Datasets.} We conduct our experiments on two popular datasets: CIFAR-10 and CIFAR-100~\cite{krizhevsky2009learning}.
A common setting is to train the model on data with equal classes in each stage \textbf{(Setting \Romannum{1})}.
Based on this, we set five stages for both CIFAR-10 and CIFAR-100, i.e., 2/20 classes in each stage. 
In addition, we further take the unequal-class scenario for different stages \textbf{(Setting \Romannum{2})}, which is more realistic in practice. 
The classes for each stage is randomly sampled and we make sure there is no overlap between different stages.  
Note that Setting \Romannum{2} is only for CIFAR-100, where the variance of class numbers is large enough for observation.

\paragraph{Implementation Details.}
All the models are implemented with PyTorch and trained on NVIDIA Tesla V100.
We use ResNet18~\cite{he2016deep} as our backbone model for experiments.
For adversarial training on both datasets, we set the maximal magnitude of perturbations $\epsilon$ to $8/255$ and  utilize the 7-step Projected Gradient Descent (PGD) to generate adversarial examples, where the step size is 2/255.
For evaluation, we not only test the standard accuracy on clean samples but also the robust accuracy with adversarial attacks.
We denote the standard accuracy as \textbf{SA}, robust accuracy under PGD attacks as \textbf{RA(PGD)}, and robust accuracy under AutoAttack as \textbf{RA(AA)}, respectively. 
The $\epsilon$ and parameters of PGD attacks for evaluation is set to be the same as for training.

During training, the class order for datasets is fixed for fair comparisons.
For reserving samples in previous stages, we use \textit{herd selection strategy} in~\cite{rebuffiICaRLIncrementalClassifier2017} and set the memory capacity to be 2000 samples for both CIFAR-10 and CIFAR-100.
The capacity is independent of the number classes and the number of exemplar for each class is $\frac{2000}{\# \text{of seen classes}}$.

\paragraph{Baselines.}
As we stated, the adversarial robustness of continual learning is firstly studied in this paper and there is no previous work on this topic.
Thus, we choose \textit{iCaRL}, the representative method for continual learning as the baseline. 
To obtain adversarial robustness, we adopt adversarial training in continual learning and build upon iCaRL, named \textit{RCL}, as another baseline.
In addition, we introduce \textit{Boundary Exemplar} to verify the influence of boundary data for \textit{RCL} 
and \textit{Mixup}, which is closely related to TABA.
\textit{Boundary Exemplar}, \textit{Mixup} and TABA are augmentation methods for improving \textit{RCL}.

\subsection{Experimental Results}
\begin{figure}[t]
\begin{minipage}[b]{.47\linewidth}
  \centering
  \centerline{\includegraphics[width=3.5cm]{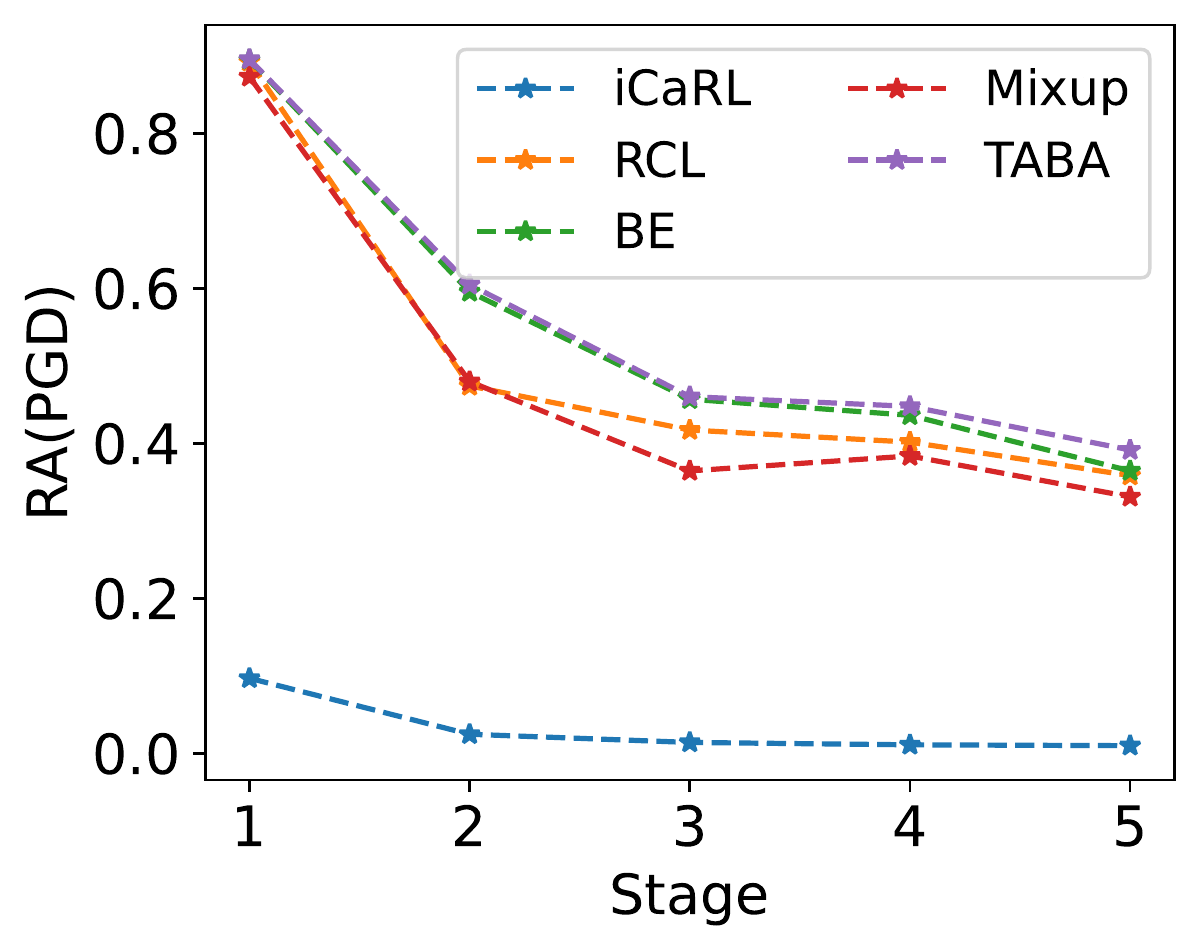}}
  \centerline{(a) CIFAR-10}\medskip
\end{minipage}
\hfill
\begin{minipage}[b]{0.47\linewidth}
  \centering
  \centerline{\includegraphics[width=3.5cm]{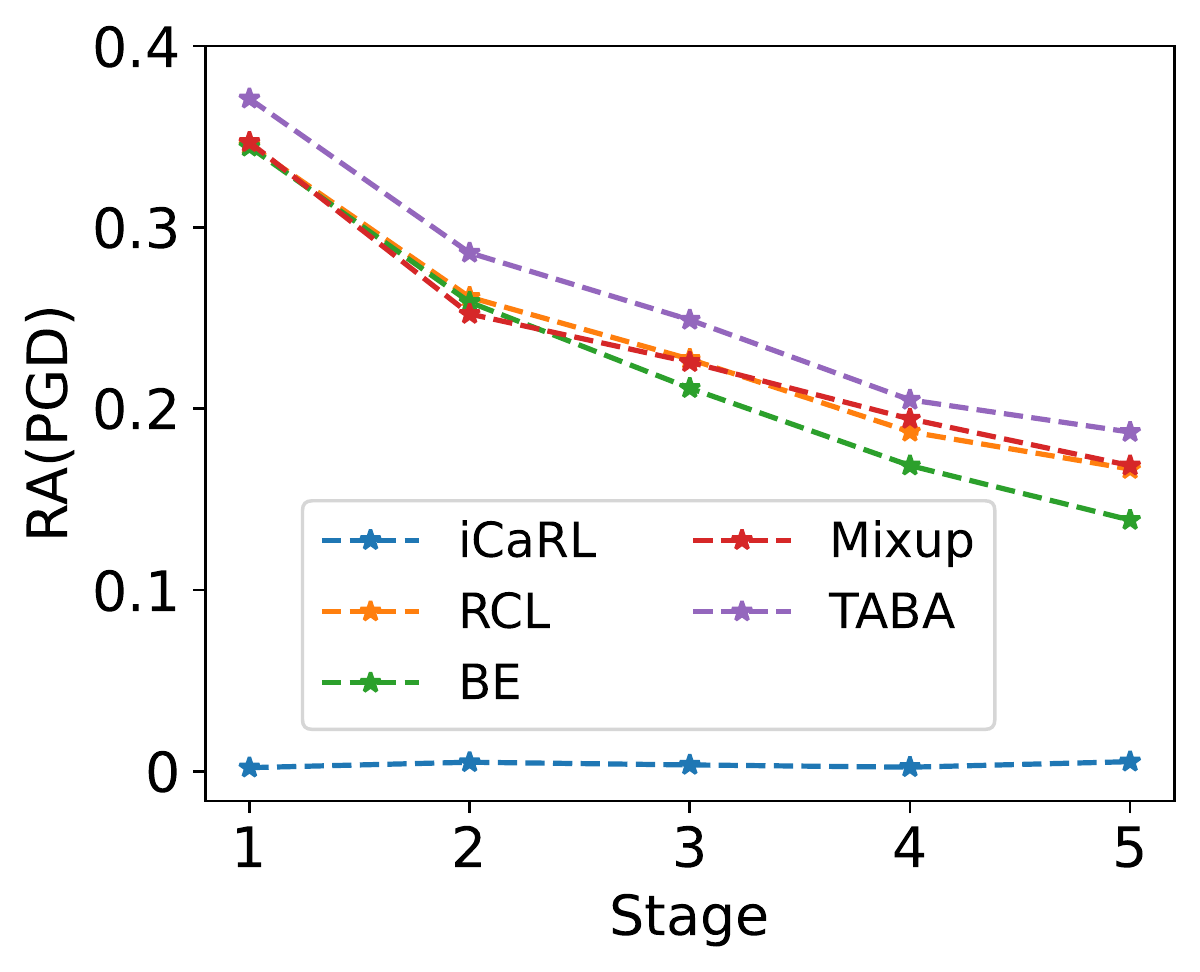}}
  \centerline{(b) CIFAR-100}\medskip
\end{minipage}
\caption{Robustness evaluation on all seen classes at different stages. BE is short for Boundary Exemplar to save space.}
\label{fig:forgetting}
\vspace{-4pt}
\end{figure}

First, we conduct experiments on CIFAR-10 and CIFAR-100 in Setting \Romannum{1}.
Robustness changes over stages are visualized in Fig.~\ref{fig:forgetting} and the experimental results are summarized in Table.~\ref{tab:setting1}.
We can observe that models trained by iCaRL are not robust %
under all %
adversarial attacks, showing nearly 0 robust accuracy against PGD attack, and 0 robust accuracy against AutoAttack.%
With adversarial training, the adversarial robustness for continual learning models is greatly improved, though there is a drop of standard accuracy.
Compared to all other methods, our TABA clearly shows strong performances: 
On both CIFAR-10 and CIFAR-100, TABA shows the best or second best robustness under PGD attacks and AutoAttacks while maintaining the standard accuracy.
Though Mixup achieves the highest robustness under AutoAttack on CIFAR10, it brings a large drop of 20\% for standard accuracy.

\begin{table}[]
\centering
\caption{Robustness evaluation on CIFAR-10 and CIFAR-100 in Setting \Romannum{1}. The best results (the higher, the better) in each column are in \textbf{bold text}.}
\label{tab:setting1}
\resizebox{0.85\columnwidth}{!}{%
\begin{tabular}{ccccc}
\toprule
\multicolumn{1}{l}{}      &                     & \textbf{SA}      & \textbf{RA(PGD)}      & \textbf{RA(AA)} \\ \midrule
\multirow{5}{*}{\rotatebox[origin=c]{90}{CIFAR10}}  & iCaRL               & \textbf{67.17\%} & 1.00\%  & 0.00\%                 \\
                          & RCL           & 60.36\% & 36.83\% & 16.71\%                \\
                          &  Boundary Exemplar & 66.52\% & 36.91\% & 10.88\%                \\
                          &  Mixup             & 46.96\% & 33.11\% & \textbf{20.36\%}                \\
                          &  TABA              & 65.97\% & \textbf{38.41\%} &  19.74\%               \\ \midrule
\multirow{5}{*}{\rotatebox[origin=c]{90}{CIFAR100}} & iCaRL               & \textbf{58.31\%} & 0.53\%  & 0.00\%                 \\
                          & RCL           & 46.67\% & 16.67\% & 9.99\%                 \\
                          &  Boundary Exemplar & 38.08\% & 14.15\% & 6.51\%                 \\
                          &  Mixup             & 46.58\% & 16.86\% & 10.03\%                \\
                          &  TABA              & 45.16\% & \textbf{18.71\%} & \textbf{11.21\%}       \\ \bottomrule        
\end{tabular}%
}
\vspace{-4pt}
\end{table}

Second, we conduct experiments in Setting \Romannum{2} on CIFAR-100.
In this setting, the class numbers for each stage are randomly selected and the sum of classes in all stages is guaranteed to be 100.
We run the experiments for 3 times and the class numbers for different stages varies from 5 to 45.
The average results are reported in Table.~\ref{tab:setting2} (the variance are close to zero and not reported here).
We can see that TABA achieves the best overall performances.
Compared to Mixup, TABA has comparable RA(AA) and much higher SA.
The large drop of SA in Mixup should be avoided.

\begin{table}[t]
\centering
\caption{Robustness evaluation on CIFAR-100 in Setting \Romannum{2}. The best results (the higher, the better) in each column are in \textbf{bold text}.}
\label{tab:setting2}
\resizebox{0.7\columnwidth}{!}{%
\begin{tabular}{cccc}
\toprule
Method    & \multicolumn{1}{c}{\textbf{SA}} & \multicolumn{1}{c}{\textbf{RA(PGD)}} & \multicolumn{1}{c}{\textbf{RA(AA)}} \\
\midrule
iCaRL     & \textbf{49.68\%}       & 0.04\%                 & 0.01\%                 \\
RCL & 44.55\%                & 17.49\%                & 9.71\%                 \\
Mixup     & 28.53\%                & 16.08\%                & \textbf{11.77\%}                \\
TABA  & 42.79\%                & \textbf{18.72\%}       & 11.43\%      \\ \bottomrule
\end{tabular}
}
\vspace{-4pt}
\end{table}

\begin{table}[t!]
\caption{Effects of three modifications in TABA.}
\label{tab:abl}
\resizebox{\columnwidth}{!}{%
\begin{tabular}{cccccc}
\toprule
\textbf{Boundary} & \textbf{Task-aware} & $\lambda$ & \textbf{SA}               & \textbf{RA(PGD)}      & \textbf{RA(AA)}                          \\ \midrule
\xmark        & \xmark & \xmark                    & 46.96\%          & 33.11\% & \multicolumn{1}{l}{\textbf{20.36\%}} \\
\cmark      & \xmark          & \xmark                    & 54.84\%          & 31.09\% & 15.45\%                     \\
\cmark       & \cmark         & \xmark                    & 59.61\%          & 32.18\% & 15.87\%                     \\
\cmark       & \cmark         & \cmark          & \textbf{65.97\%} & \textbf{38.41\%} & 19.74\%    \\ \bottomrule                
\end{tabular}
}
\footnotesize{\cmark: w/ \quad \xmark: w/o}\\
\vspace{-4pt}
\end{table}

\subsection{Ablation Study}
Inspired by Mixup, we propose TABA for relieving the forgetting of adversarial robustness in continual learning.
Compared to Mixup, TABA is different in three ways: \textit{boundary data}, \textit{task-aware sample selection} and \textit{the range of} $\lambda$.
Here we investigate the effects of these modifications and results are summarized in Table.~\ref{tab:abl}.
We can see the improvements when we make modifications sequentially on Mixup.

\section{Conclusion}
\label{sec:con}
In this paper, we study the continual learning problem in the adversarial settings. 
It is verified that models trained in continual learning ways are also vulnerable to adversarial examples.
We thus propose RCL-TABA, which consists of adversarial training and a novel data augmentation method TABA, to secure continual learning. 
As this is the very first step to studying the intersection of adversarial training and continual learning, we hope our findings provide useful insights and motivate researchers to explore deeper.

\bibliographystyle{IEEEbib}
\bibliography{strings,refs}

\end{document}